%% file: acmmm16.tex
\documentclass[10pt,twocolumn,letterpaper]{article}

\usepackage{cvpr}

\usepackage{tikz}
\usetikzlibrary{shapes,arrows}
\usetikzlibrary{positioning}
\usetikzlibrary{shapes.misc}
\usepackage{pgfplots}
\usepackage{tabularx}
\usepackage{hyperref}
\usepackage{booktabs}
\usepackage{subcaption}
\usepackage{todonotes}
\usepackage{soul}
\usepackage{url}
\usepackage{amsmath}
\usepackage{amssymb}

\hypersetup{%
  pdftitle={Visual Congruent Ads for Image Search},
 pdfauthor={Yannis Kalantidis, Ayman Farahat, Lyndon Kennedy, Ricardo Baeza-Yates, David A. Shamma},
  pdfkeywords={online advertising, visual similarity, ad selection, native ads, image search},
  bookmarksnumbered,
  pdfstartview={FitH},
  colorlinks,
  citecolor=black,
  urlcolor=black,
  filecolor=black,
  linkcolor=black,
  breaklinks=true,
}
\cvprfinalcopy % *** Uncomment this line for the final submission

\def\cvprPaperID{****} % *** Enter the CVPR Paper ID here

\begin{document}
% \sloppy

\title{Visual Congruent Ads for Image Search}

%--------------------------------------------------------------------
% Authors
\author{Yannis Kalantidis\thanks{Corresponding author, email: ykal@yahoo-inc.com} , Ayman Farahat\thanks{Work done while authors were at Yahoo Labs} , Lyndon Kennedy$^\dagger$, Ricardo Baeza-Yates$^\dagger$ and David A. Shamma$^\dagger$ \\ 
Yahoo Inc. \\
San Francisco, CA
}
%--------------------------------------------------------------------

\input{tex/abbrev}

\maketitle
\begin{abstract}
The quality of user experience online is affected by the relevance and placement of advertisements. We propose a new system for selecting and displaying visual advertisements in image search result sets. Our method compares the visual similarity of candidate ads to the image search results and selects the most visually similar ad to be displayed. The method further selects an appropriate location in the displayed image grid to minimize the perceptual visual differences between the ad and its neighbors. We conduct an experiment with about 900 users and find that our proposed method provides significant improvement in the users' overall satisfaction with the image search experience,  without diminishing the users' ability to see the ad or recall the advertised brand.
\end{abstract}

% introduction

\input{tex/intro}

% related work
% \vspace{.5cm}
\input{tex/related}

% main section
% \vspace{.5cm}
\input{tex/method}

% experiments
\input{tex/experiments}

% discussion
% \vspace{.5cm}
\input{tex/discussion}

% conclusions
% \vspace{.5cm}
\input{tex/conclusions}

% references
% \newpage
% \clearpage
\bibliographystyle{abbrv}
\bibliography{tex/bib}  % sigproc.bib is the name of the Bibliography in this case
%

% appendix
% \vspace{.5cm}
% \input{tex/appendix}

% iiiiffff and when it get accepted, thank Sanjay

\end{document}

%% file: tex/abbrev.tex
\newcommand{\nospace}{\vspace{-4mm}}
\newcommand{\head}[1]{{\smallskip\noindent\bf #1}}
\newcommand{\alert}[1]{\hl{#1}}
\newcommand{\suppl}{\emph{given in the supplementary material}}

%--------------------------------------------------------------------------------

\newcommand{\one}{\mathbbm{1}}
\newcommand{\bbF}{\mathbb{F}}
\newcommand{\bbH}{\mathbb{H}}
\newcommand{\expect}{\mathbb{E}}
\newcommand{\project}{\mathbb{P}}
\newcommand{\real}{\mathbb{R}}
\newcommand{\tran}{^{\mathrm{T}}}
\newcommand{\diff}{\mathrm{d}}
\newcommand{\prob}{\mathrm{Pr}}
\newcommand{\binomial}{\mathrm{Bi}}
\newcommand{\weibull}{\mathrm{Wb}}
\newcommand{\normal}{\mathcal{N}}
\newcommand{\wishart}{\mathcal{W}}
\newcommand{\zcol}{\mathbf{0}}
\newcommand{\zrow}{\zcol\tran}
\newcommand{\otherwise}{\mathrm{otherwise}}

\newcommand{\cA}{\mathcal{A}}
\newcommand{\cB}{\mathcal{B}}
\newcommand{\cC}{\mathcal{C}}
\newcommand{\cD}{\mathcal{D}}
\newcommand{\cE}{\mathcal{E}}
\newcommand{\cF}{\mathcal{F}}
\newcommand{\cG}{\mathcal{G}}
\newcommand{\cH}{\mathcal{H}}
\newcommand{\cI}{\mathcal{I}}
\newcommand{\cJ}{\mathcal{J}}
\newcommand{\cK}{\mathcal{K}}
\newcommand{\cL}{\mathcal{L}}
\newcommand{\cM}{\mathcal{M}}
\newcommand{\cN}{\mathcal{N}}
\newcommand{\cO}{\mathcal{O}}
\newcommand{\cP}{\mathcal{P}}
\newcommand{\cQ}{\mathcal{Q}}
\newcommand{\cR}{\mathcal{R}}
\newcommand{\cS}{\mathcal{S}}
\newcommand{\cT}{\mathcal{T}}
\newcommand{\cU}{\mathcal{U}}
\newcommand{\cV}{\mathcal{V}}
\newcommand{\cW}{\mathcal{W}}
\newcommand{\cX}{\mathcal{X}}
\newcommand{\cY}{\mathcal{Y}}
\newcommand{\cZ}{\mathcal{Z}}

\newcommand{\cII}{\mathcal{I}_f}
\newcommand{\cAA}{\mathcal{A}_f}
\newcommand{\It}{i}
\newcommand{\At}{a}

\newcommand{\va}{\mathbf{a}}
\newcommand{\vb}{\mathbf{b}}
\newcommand{\vc}{\mathbf{c}}
\newcommand{\vd}{\mathbf{d}}
\newcommand{\ve}{\mathbf{e}}
\newcommand{\vf}{\mathbf{f}}
\newcommand{\vg}{\mathbf{g}}
\newcommand{\vh}{\mathbf{h}}
\newcommand{\vi}{\mathbf{i}}
\newcommand{\vj}{\mathbf{j}}
\newcommand{\vk}{\mathbf{k}}
\newcommand{\vl}{\mathbf{l}}
\newcommand{\vm}{\mathbf{m}}
\newcommand{\vn}{\mathbf{n}}
\newcommand{\vo}{\mathbf{o}}
\newcommand{\vp}{\mathbf{p}}
\newcommand{\vq}{\mathbf{q}}
\newcommand{\vr}{\mathbf{r}}
\newcommand{\vt}{\mathbf{t}}
\newcommand{\vu}{\mathbf{u}}
\newcommand{\vv}{\mathbf{v}}
\newcommand{\vw}{\mathbf{w}}
\newcommand{\vx}{\mathbf{x}}
\newcommand{\vy}{\mathbf{y}}
\newcommand{\vz}{\mathbf{z}}

\newcommand{\vA}{\mathbf{A}}
\newcommand{\vB}{\mathbf{B}}
\newcommand{\vC}{\mathbf{C}}
\newcommand{\vD}{\mathbf{D}}
\newcommand{\vF}{\mathbf{F}}
\newcommand{\vG}{\mathbf{G}}
\newcommand{\vI}{\mathbf{I}}
\newcommand{\vM}{\mathbf{M}}
\newcommand{\vS}{\mathbf{S}}
\newcommand{\vW}{\mathbf{W}}
\newcommand{\vX}{\mathbf{X}}
\newcommand{\vzero}{\mathbf{0}}

\newcommand{\rLambda}{\mathrm{\Lambda}}
\newcommand{\rSigma}{\mathrm{\Sigma}}

\newcommand{\vmu}{\bm{\mu}}
\newcommand{\vpi}{\bm{\pi}}
\newcommand{\vLambda}{\bm{\rLambda}}
\newcommand{\vSigma}{\bm{\rSigma}}

%--------------------------------------------------------------------------------
% long/short versions and change tracking

\newcommand{\rep}[2]{%
   \ifthenelse{\equal{\version}{track}}%
      {{\color{gray}%\sout % \sout not working in math mode!
{#1}}{\color{blue}{#2}}}% needs ulem package
      {\ifthenelse{\equal{\version}{long}}{#1}{#2}}%
}

\newcommand{\cut}[1]{\rep{#1}{}}
\newcommand{\ins}[1]{\rep{}{#1}}

% example:

% \noindent
% \emph{Cut a part of the text:}
%
% \noindent
% \def\version{short}%
% This is long\cut{er and longer}.\\
% \def\version{long}%
% This is long\cut{er and longer}.\\
% \def\version{track}%
% This is long\cut{er and longer}.\\
%
% \noindent
% \emph{Replace part of the text with a shorter version:}
%
% \noindent
% \def\version{short}%
% This is long\rep{er and longer}{ enough}.\\
% \def\version{long}%
% This is long\rep{er and longer}{ enough}.\\
% \def\version{track}%
% This is long\rep{er and longer}{ enough}.\\

%--------------------------------------------------------------------------------
% Add a period to the end of an abbreviation unless there's one
% already, then \xspace.
% \makeatletter
% \DeclareRobustCommand\onedot{\futurelet\@let@token\@onedot}
% \def\@onedot{\ifx\@let@token.\else.\null\fi\xspace}
% \def\onedot{.~}
\def\onedot{.~}

\def\eg{\emph{e.g}.,~} \def\Eg{\emph{E.g}\onedot}
\def\ie{\emph{i.e}.,~} \def\Ie{\emph{I.e}\onedot}
\def\cf{\emph{c.f}\onedot} \def\Cf{\emph{C.f}\onedot}
\def\etc{\emph{etc}} %\def\vs{\emph{vs}\onedot}
\def\wrt{w.r.t\onedot} \def\dof{d.o.f\onedot}
\def\etal{\emph{et al}\onedot}
\makeatother

%% file: tex/intro.tex
\section{Introduction}
\label{sec:intro}
\vspace{.2cm}

\begin{figure}[t!]
\centering
 	\includegraphics[width=\columnwidth]{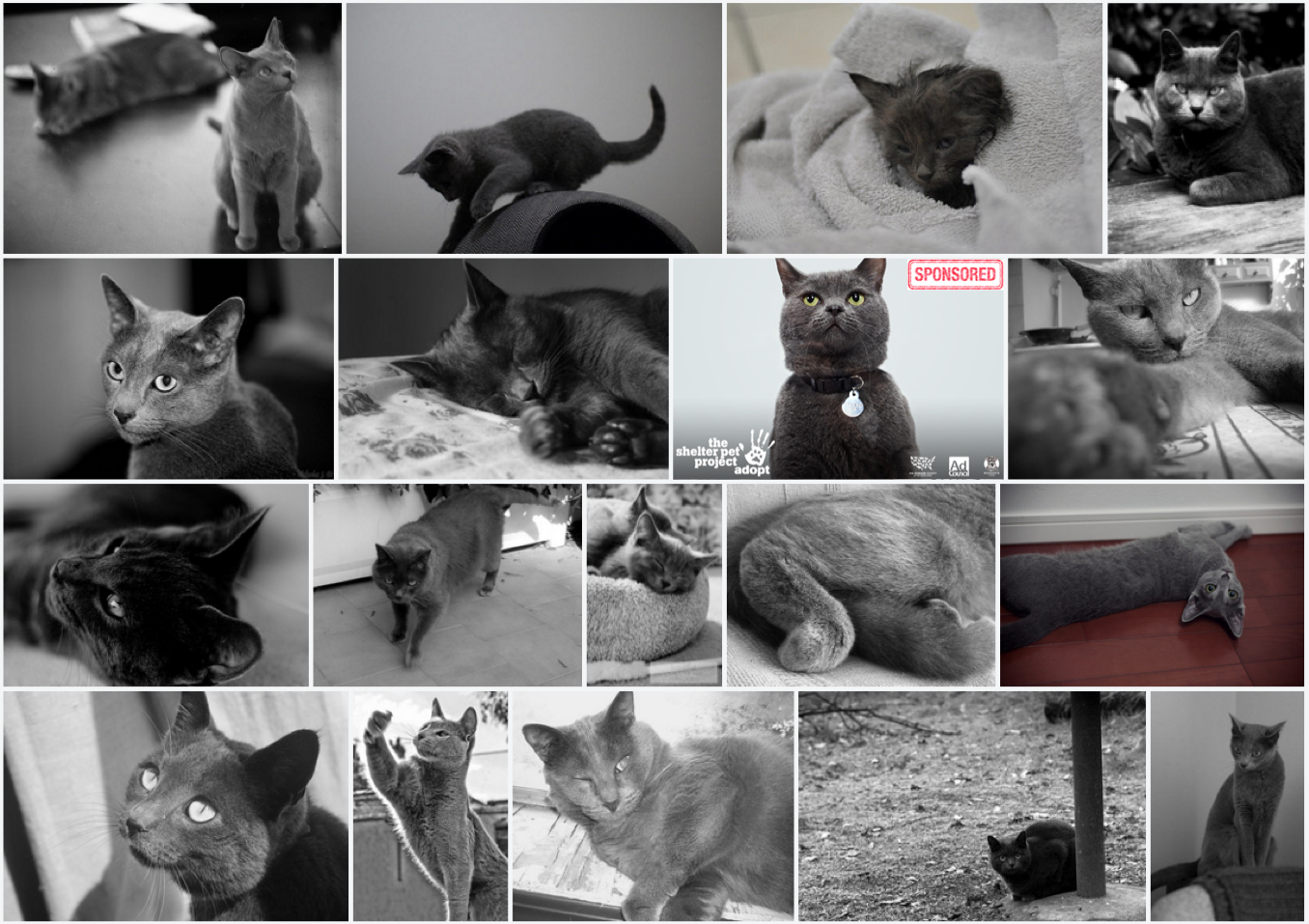}
    \caption{An example of visually congruent ad placement in image search. The ad matches both visually and semantically with the surrounding results.}
	\label{fig:front}
\end{figure}

\begin{figure*}[t]
\centering
	\input{fig/overview}
    \caption{Overview of the proposed approach. Advertisement images are shown in gray squares, while the one that is subsequently selected for display is shown as a blue square with a red border and a cross. Squares filled with any other color correspond to images from the same cluster in visual space.}
	\label{fig:overview}
\end{figure*}
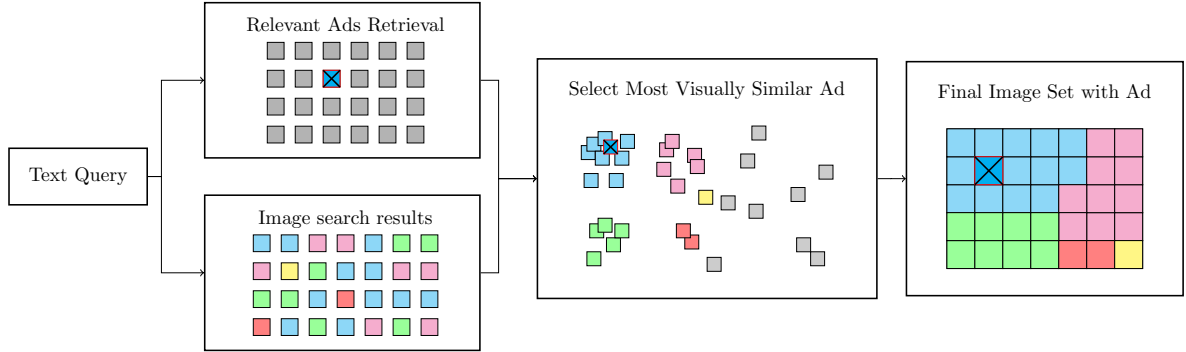

\noindent
The goal of native advertising is to present sponsored content in a way that matches the surrounding context in which the advertisements appear. Displaying ads online generally allows advertisers to target users with lower cost than in offline settings~\cite{Gold14}. Furthermore, the additional information about site content and user interests available to online advertisers allows them to choose ads which are particularly semantically relevant and, thus, appealing or memorable for users. For these reasons, native advertisements have rightfully taken over a large chunk of the market~\cite{GoTu11a}.

While newer to digital advertising, the underlying concept of native advertisements has been well established in offline domains, such as print magazines. \textit{Vanity Fair}, a magazine covering topics in fashion, popular culture, and entertainment, has used an in-house design creative agency (Vanity Fair Agenda) since 2003 to ensure that ads match the look and feel of the surrounding content~\cite{vf:ads}. An ideal magazine ad creates a memorable experience for a brand while becoming a part of the experience of reading the magazine, rather than distracting from it.

It is estimated that $25\%$ of search queries have image-related intent. Effective placement of native advertisements within image search results has been explored far less than in the context of traditional search. Thus, techniques which optimize the placement of native image ads present a significant opportunity. Optimizing image results in the context of photo-oriented sites or image search engines presents an interesting tradeoff. Image queries are very difficult to monetize, meaning that presenting ads that are relevant and clearly marked can lead to significant increases in revenue. On the other hand, advertisements which add clutter or distract from organic results can degrade the overall search experience. Advertisers must find a way to make their content stand out while still matching the surrounding content.

%Native advertising presents several unique opportunitines for presenting advertisements to users at lower cost and with higher relevance. First, ads can be displayed at lower costs in online contexts than offline~\cite{Gold14}. Furthermore, by 
%Display advertising is the driving force behind the growing Internet market. As the cost of targeted advertising online is much lower~\cite{Gold14}, \emph{native} advertisements have rightfully taken over a big chunk of the market~\cite{GoTu11a}. This is accompanied by a trend in which advertisers are pursuing ads which are increasingly semantically relevant to both surrounding context and the interests of the specific user. Optimizing image results from photo-oriented sites or image search engines presents an interesting problem. Many images in a grid layout can lead to advertisements (ads) adding clutter and degrading the overall search experience. However image queries are very hard to monetize, hence any improvements on how ads are placed in image search results may have a good revenue impact; it is estimated that $25\%$ of search queries have images intent. Most advertisement placing spaces on the Web have been explored, placing advertisements in image search is not well explored.
 
Prior research has shown that finding the appropriate balance can be especially difficult in contexts in which users are browsing rich multimedia content. Increasing the obtrusiveness of multimedia-based advertisements has been shown to increase purchase intent~\cite{CoSF12}. However, making ads more obtrusive runs the risk of alienating users and creating a negative impression of both the site publisher and the brand being advertised~\cite{Gold14}. The combination of relevance to website content and advertisement obtrusiveness has also been shown to raise additional privacy concerns~\cite{Gotu11b}, further complicating matters. The quality of user experience is highly affected by the quality, relevance and placement of advertisements in a webpage~\cite{AZZ+12,dSVC13}.

% In this paper, we focus on the problem of serving image-based advertisements with no animation. Although images are still more attention-grabbing than text-based ads, we believe that high-quality image-based advertisements can be chosen in a manner that is not distracting or annoying to users. 
We focus on the selection and placement of high-quality image-based advertisements \emph{with an emphasis on not distracting or annoying to users}.
Ideally, we want to place an ad that is properly marked and that blends in with the rest of the image set, adding to the overall user experience rather than detracting from it, as in Figure~\ref{fig:front}. Although we want the advertisement to blend in with the surrounding context, it is also important that it be clearly marked. Deceiving or tricking the user with a camouflaged advertisement can degrade user trust, harming the site in the long run.

Therefore, we propose a specific technique that strikes a balance between making users aware of a native image ad without distracting from the overall image search experience. We accomplish this by going one step further than semantic similarity between advertisement and web content; we propose using \emph{visual similarity} between image-based ads and surrounding images to increase their quality and effectiveness. We demonstrate that this approach makes native image ads seem more relevant and less distracting to users, without hurting clicks or recognition of the ad.

An overview of our approach is presented in Figure~\ref{fig:overview} and summarized here. Starting with a set of image search results and a set of relevant advertisements, we first select the advertisement most visually similar to the result set; We then position it within the results, either preserving the initial ordering of results or reorganizing the whole set based on visual similarity. We refer to this approach as \emph{Visually Congruent Ad Placement} or \emph{VCAP}.

We can summarize our contributions as follows:
\begin{enumerate}
\item We propose the use of visual similarity between ad images and image search results and present a simple, automated way of selecting visually congruent native ads. We suggest different ways of positioning the ad in the result set grid in a way that the ad image is displayed among its most visually similar ones. 
\item We use compressed image features to make our approach applicable to web-scale image search. Our algorithm runs in a few hundred milliseconds and requires storing only a few bytes per image.
\item We present findings from a large crowd sourced user study that shows that our algorithm has a sizable and statistically significant impact on reducing user distraction and increasing their overall ad experience while not hurting ad recognition. Given the absence of ad revenue from image search, our findings have significant financial implications.
\end{enumerate}

Our experimental validation on about $900$ human samples through Amazon Mechanical Turk\footnote{\url{https://www.mturk.com/}} shows that our approach is able to retain brand recognition and awareness as effectively as a ``distracting'' ad would, while at the same time providing for a better user experience.

To the best of our knowledge, this is the first paper that investigates and evaluates visually congruent native advertisements in image search. The paper is structured as follows:  \S~\ref{sec:related} first discusses related work. \S~\ref{sec:method} presents the core of our approach, \ie selection and positioning of the ad in a web-scale engine. \S~\ref{sec:experiments} presents our experimental validation while  \S~\ref{sec:discussion} discusses different aspects, generalizations and implications of our approach. The paper concludes in \S~\ref{sec:conclusions}.

%% file: fig/overview.tex
\centering
\resizebox{.9\linewidth}{!}{%
\begin{tikzpicture}[scale=.08,
    block_center/.style ={rectangle, draw=black, thick, fill=white,
      text centered, inner sep=10pt},
    block_center2/.style ={rectangle, draw=black, thick, fill=white,
      text centered, minimum width=14em, inner sep=2pt},
	block_left/.style ={rectangle, draw=black, thick, fill=white,
      text ragged, minimum height=4em, inner sep=10pt},
      ]

	% --------------------------------------------------------------------
	% NODE text query
	\node [block_center] (referred) {Text Query};
	% --------------------------------------------------------------------

	% --------------------------------------------------------------------
	% NODE Relevant Ads Retrieval
	\node [block_center2, above right= -.2cm and 1cm of referred] (excluded1) {%
		\def\arraystretch{1.5}
        \begin{tabular}{c}
          Relevant Ads Retrieval \\ 
          \begin{tikzpicture}[scale=.5]
              \foreach \x in {0,...,5}
                  \foreach \y in {0,...,3} 
                      \filldraw[black!30!white, draw=black] (\x,\y) rectangle (\x+.6,\y + 0.6);
              \filldraw[cyan, draw=red] (2,2) rectangle (2.6,2.6);
              \draw[black,thick] (2,2) -- (2.6,2.6);
              \draw[black,thick] (2.6,2) -- (2,2.6);
          \end{tikzpicture} \\ 
		\end{tabular}
    };
	% --------------------------------------------------------------------
	
	% --------------------------------------------------------------------
	% NODE Image search results
    \node [block_center2, below right = -.2cm and 1cm of referred] (excluded2) {%
		\def\arraystretch{1.5}
		\begin{tabular}{c}
			Image search results \\
			\begin{tikzpicture}[scale=.5]
% 				\foreach \x in {0,...,6}
% 					\foreach \y in {0,...,3}
% 				\filldraw[white, draw=black] (\x,\y) rectangle (\x+.6,\y + .6);
                \filldraw[red!50!white, draw=black] ( 0 , 0 ) rectangle ( 0.6 , 0.6 );
                \filldraw[green!40!white, draw=black] ( 0 , 1 ) rectangle ( 0.6 , 1.6 );
                \filldraw[magenta!40!white, draw=black] ( 0 , 2 ) rectangle ( 0.6 , 2.6 );
                \filldraw[cyan!40!white, draw=black] ( 0 , 3 ) rectangle ( 0.6 , 3.6 );
                \filldraw[cyan!40!white, draw=black] ( 1 , 0 ) rectangle ( 1.6 , 0.6 );
                \filldraw[green!40!white, draw=black] ( 1 , 1 ) rectangle ( 1.6 , 1.6 );
                \filldraw[yellow!60!white, draw=black] ( 1 , 2 ) rectangle ( 1.6 , 2.6 );
                \filldraw[cyan!40!white, draw=black] ( 1 , 3 ) rectangle ( 1.6 , 3.6 );
                \filldraw[green!40!white, draw=black] ( 2 , 0 ) rectangle ( 2.6 , 0.6 );
                \filldraw[cyan!40!white, draw=black] ( 2 , 1 ) rectangle ( 2.6 , 1.6 );
                \filldraw[green!40!white, draw=black] ( 2 , 2 ) rectangle ( 2.6 , 2.6 );
                \filldraw[magenta!40!white, draw=black] ( 2 , 3 ) rectangle ( 2.6 , 3.6 );
                \filldraw[cyan!40!white, draw=black] ( 3 , 0 ) rectangle ( 3.6 , 0.6 );
                \filldraw[red!50!white, draw=black] ( 3 , 1 ) rectangle ( 3.6 , 1.6 );
                \filldraw[cyan!40!white, draw=black] ( 3 , 2 ) rectangle ( 3.6 , 2.6 );
                \filldraw[magenta!40!white, draw=black] ( 3 , 3 ) rectangle ( 3.6 , 3.6 );
                \filldraw[magenta!40!white, draw=black] ( 4 , 0 ) rectangle ( 4.6 , 0.6 );
                \filldraw[cyan!40!white, draw=black] ( 4 , 1 ) rectangle ( 4.6 , 1.6 );
                \filldraw[cyan!40!white, draw=black] ( 4 , 2 ) rectangle ( 4.6 , 2.6 );
                \filldraw[cyan!40!white, draw=black] ( 4 , 3 ) rectangle ( 4.6 , 3.6 );
                \filldraw[green!40!white, draw=black] ( 5 , 0 ) rectangle ( 5.6 , 0.6 );
                \filldraw[cyan!40!white, draw=black] ( 5 , 1 ) rectangle ( 5.6 , 1.6 );
                \filldraw[magenta!40!white, draw=black] ( 5 , 2 ) rectangle ( 5.6 , 2.6 );
                \filldraw[green!40!white, draw=black] ( 5 , 3 ) rectangle ( 5.6 , 3.6 );
                \filldraw[magenta!40!white, draw=black] ( 6 , 0 ) rectangle ( 6.6 , 0.6 );
                \filldraw[cyan!40!white, draw=black] ( 6 , 1 ) rectangle ( 6.6 , 1.6 );
                \filldraw[magenta!40!white, draw=black] ( 6 , 2 ) rectangle ( 6.6 , 2.6 );
                \filldraw[green!40!white, draw=black] ( 6 , 3 ) rectangle ( 6.6 , 3.6 );
			\end{tikzpicture}
        \end{tabular}
	};
	% --------------------------------------------------------------------
    
	% --------------------------------------------------------------------
	% NODE Select Most Visually Similar Ad
	\node [block_left, below right= -1.8cm and 1cm of excluded1] (excluded3) {%
		\begin{tabular}{c}
			Select Most Visually Similar Ad \\ \\ 
			\begin{tikzpicture}[scale=.5]
%				%These used to be placeholders....            	
% 				\foreach \x in {0,...,6}
% 					\foreach \y in {0,...,4}
% 						\draw[white] (\x,\y) rectangle (\x+1,\y + 1);

				% ----------------------------
				% Cyan/ad cluster
                \filldraw[cyan!40!white, draw=black] (2.3,3) rectangle (2.8,3.5);
                \filldraw[cyan!40!white, draw=black] (1.4,3) rectangle (1.9,3.5);
                \filldraw[cyan!40!white, draw=black] (1.3,4) rectangle (1.8,4.5);
                \filldraw[cyan!40!white, draw=black] (1.5,4.3) rectangle (2,4.8);
                \filldraw[cyan!40!white, draw=black] (1.9,4.5) rectangle (2.4,5);
                \filldraw[cyan!40!white, draw=black] (2.7,4.4) rectangle (3.2,4.9);
                \filldraw[cyan!40!white, draw=black] (2.5,3.8) rectangle (3,4.3);
                \filldraw[cyan!40!white, draw=black] (1.8,3.8) rectangle (2.3,4.3);
                \filldraw[cyan, draw=red] (2.1,4.2) rectangle (2.6,4.7);
				\draw[black,thick] (2.1,4.2) -- (2.6,4.7);
              	\draw[black,thick] (2.1,4.7) -- (2.6,4.2);
				% pink cluster
                \filldraw[magenta!40!white, draw=black] (4.1,4.1) rectangle (4.6,4.6);
                \filldraw[magenta!40!white, draw=black] (4.5,2.8) rectangle (5,3.3);
                \filldraw[magenta!40!white, draw=black] (4,3.4) rectangle (4.5,3.9);
                \filldraw[magenta!40!white, draw=black] (4.3,4.4) rectangle (4.8,4.9);
                \filldraw[magenta!40!white, draw=black] (5.1,3.9) rectangle (5.6,4.4);
                \filldraw[magenta!40!white, draw=black] (5.2,3.5) rectangle (5.7,4);

				% yellow cluster
				\filldraw[yellow!60!white, draw=black] (5.5,2.4) rectangle (6,2.9);

				% red cluster
                \filldraw[red!50!white, draw=black] (5.0,0.8) rectangle (5.5,1.3);
                \filldraw[red!50!white, draw=black] (4.7,1.2) rectangle (5.2,1.7);

				% green cluster
                \filldraw[green!40!white, draw=black] (1.6,1.2) rectangle (2.1,1.7);
                \filldraw[green!40!white, draw=black] (2.2,0.7) rectangle (2.7,1.2);
                \filldraw[green!40!white, draw=black] (1.9,1.4) rectangle (2.4,1.9);
                \filldraw[green!40!white, draw=black] (2.5,1.2) rectangle (3,1.7);
                \filldraw[green!40!white, draw=black] (1.5,0.2) rectangle (2,0.7);

\filldraw[black!20!white, draw=black] (9.5,0.2) rectangle (10,0.7);
\filldraw[black!20!white, draw=black] (8.8,2.5) rectangle (9.3,3);
\filldraw[black!20!white, draw=black] (9.8,3.3) rectangle (10.3,3.8);

\filldraw[black!20!white, draw=black] (9,0.7) rectangle (9.5,1.2);
\filldraw[black!20!white, draw=black] (7,3.7) rectangle (7.5,4.2);
\filldraw[black!20!white, draw=black] (7.4,4.7) rectangle (7.9,5.2);
\filldraw[black!20!white, draw=black] (6.3,2.2) rectangle (6.8,2.7);
\filldraw[black!20!white, draw=black] (7.3,1.9) rectangle (7.8,2.4);
\filldraw[black!20!white, draw=black] (5.8,0) rectangle (6.3,0.5);

\end{tikzpicture}       
              		\end{tabular}

      }; 
      \node [block_left, right=.5cm of excluded3] (random) {%
         	\begin{tabular}{c}
		Final Image Set with Ad\\ \\ 
        
      \begin{tikzpicture}[scale=.5]
\foreach \x in {0,...,6}
	\foreach \y in {0,...,4}
		\draw (\x,\y) rectangle (\x+1,\y + 1);
        
\foreach \x in {0,...,4}
	\foreach \y in {2,...,4}
		\filldraw[cyan!40!white, draw=black](\x,\y) rectangle (\x+1,\y + 1);
\filldraw[cyan, draw=red] (1,3) rectangle (2,4);
              \draw[black,thick] (1,3) -- (2,4);
              \draw[black,thick] (1,4) -- (2,3);
\foreach \x in {5,...,6}
	\foreach \y in {1,...,4}
		\filldraw[magenta!40!white, draw=black](\x,\y) rectangle (\x+1,\y + 1);
\filldraw[magenta!40!white, draw=black](4,2) rectangle (5,3);
\filldraw[magenta!40!white, draw=black](4,1) rectangle (5,2);
\foreach \x in {0,...,3}
	\foreach \y in {0,...,1}
		\filldraw[green!40!white, draw=black](\x,\y) rectangle (\x+1,\y + 1);
\filldraw[red!50!white, draw=black](4,0) rectangle (5,1);
\filldraw[red!50!white, draw=black](5,0) rectangle (6,1);
\filldraw[yellow!60!white, draw=black](6,0) rectangle (7,1);

\end{tikzpicture} 
\end{tabular}

      };

		\draw[->] (referred.east) -- ++(3,0) |- (excluded1.west);
      	\draw[->] (referred.east) -- ++(3,0) |- (excluded2.west);
      	\draw[->] (excluded1.east) -- ++(4,0) |- (excluded3.west);
      	\draw[->] (excluded2.east) -- ++(4,0) |- (excluded3.west);
      	\draw[->] (excluded3.east) -- ++(3,0) |- (random.west);

\end{tikzpicture}
}

%% file: tex/related.tex
\vspace{.2cm}
\section{Related work}
\label{sec:related}
\vspace{.2cm}

%\head{Image content analysis \& description}.

Image content representation has evolved greatly during the last years. The first wave started with the success of local SIFT features~\cite{Lowe04} and the bag-of-words model for search~\cite{SiZi03}, borrowed from text retrieval~\cite{BYRN11}. As the field evolved, focus shifted on global image representations for efficiency, through aggregating local features. Most notable examples of aggregation are VLAD~\cite{JDSP10} and Fisher Vectors~\cite{PLSP10}, two approaches that have influenced multiple 
% extensions~\cite{ToAJ13,DGJP13,GMJP14,ToJA15}.
extensions~\cite{ToAJ13,DGJP13,GMJP14}.

We are currently in a new era for image content representation. With the recent advances in GPUs and the growing number of training data available, the computer vision community has revisited Deep Convolutional Neural Networks (CNNs), \ie neural nets with many hidden layers and millions of parameters. When trained on big image databases like ImageNet~\cite{RDS+14}, CNNs have been shown to ``effortlessly'' improve the previous state-of-the-art in many computer vision applications, \eg image classification~\cite{KrSH12} and visual search~\cite{BSCL14}. 

The aggregated approaches as well as the CNN-based ones, all produce a \emph{global} image signature, \ie a high-dimensional feature vector in Euclidean space. In this case, similarity search reduces to nearest neighbor search in feature space allowing for very scalable search. When the image database consists of billions of images, every byte counts. We therefore need to embed our visual features to a very compact image signature for storage, that also keeps nearest neighbor search performance close to the original space. 

Large scale nearest neighbor search was traditionally based on hashing signatures~\cite{DIIM04, PaJA10}, mainly due to low memory footprints and fast search in Hamming space~\cite{NoPF12}. But even recent advances in the field~\cite{JHL+13}, cannot achieve performance comparable to the slower but still memory efficient quantization-based approaches~\cite{JeDS11,NoPF12}. What is more important, approaches that are based on Product Quantization, unlike hashing, allow us to reconstruct the original vectors from the compact signatures. The past approaches~\cite{JeDS11, GHKS14, KaAv14} are all good candidates for compressing CNN visual features. We choose to use the Locally Optimized Product Quantization approach~\cite{KaAv14} that currently gives the state-of-the-art results in nearest neighbor search.  

% Our work directly addresses the issues of advertising externalities and image search monetization. While advertising is the main revenue driver on the Internet, little is known about its impact on user utility. Goldstein \etal\cite{GoMS13} conducted a crowd-sourcing experiment to put a monetary value on the cost of annoying ads. Users were given the opportunity to categorize emails for a per-message wage and quit at any time. Participants were randomly assigned to one of three different pay rates and also randomly assigned to categorize the emails in the presence of no ads, annoying ads, or innocuous ads. Using a wage differential approach of Tommim \etal\cite{TKPL11}, the authors found that users who were exposed to bad ads needed to get an additional pay raise of $\$1.15$ for every 1000 impressions they generated.
% De Sa \etal\cite{dSNC13} also confirmed these finding for display ads on mobile devices. It was also recently shown that the increase in TV advertising from 6 minutes an hour to the current 14 minutes has been a major driver of DVR adoption~\cite{TeWP10}.

% To the best of our knowledge, there has been no work that directly attempts to quantify the externalities of ad placement on image search.

%% file: tex/method.tex
\section{VCAP: Visually Congruent Ad Placement}
\label{sec:method}
\vspace{.2cm}

\noindent
Let $\cI = \{ I_1, \ldots, I_n\}$ be the set of $n$ images returned as search results for a text query $q$ by user $u$. Also let $\cA = \{ A_1, \ldots, A_m\}$ be a set of $m$ image-based \emph{advertisements} that are  \emph{relevant} to the query and/or the user.  An advertisement can be relevant to the query either in terms of semantic/topical similarity with $q$ (\eg a car advertisement for the query ``convertible cars'') or in terms of the interests of the user $u$, \ie based on any profile or browsing history data that might be available (\eg a car advertisement for a user that frequents car-related sites). 

Selecting a relevant or semantically similar ad is a subject that has been studied extensively~\cite{rel-ads}. Here, we assume that the \emph{set} $\cA$ of such relevant ads is already available to us for query $q$. What we aim for is to automatically select the advertisement from set $\cA$ that best matches the image set $\cI$ in terms of visual appearance and also present/place it in such a way that it fits seamlessly with its surrounding images. 
% This section continues as follows: We first present the features we use for representing the visual content of images in \S~\ref{subsec:visfeat} and our ad selection approach in \S~\ref{subsec:selection}. We then present ways of positioning the ad inside the image grid in \S~\ref{subsec:place} and finally propose a scalable approach for real-time and web-scale application of our algorithm in \S~\ref{subsec:scaling}.

% Ad selection
% \vspace{.5cm}
\input{tex/method_selection}

% ad placement
% \vspace{.5cm}
\input{tex/method_positioning}

% large-scale with LOPQ
% \vspace{.5cm}
\input{tex/method_scaling}

%% file: tex/method_selection.tex
% \subsection{Selecting the Most Visually Similar Ad}
% \label{subsec:selection}
% \vspace{.2cm}

% \subsection{Compact visual features}
% \label{subsec:visfeat}
% \vspace{.2cm}

% % \head{Compact visual features}. 
% \noindent
Our approach is based on \emph{visual similarity}, \ie similarity based on image visual content; here we choose features extracted by Deep Convolutional Neural Networks (CNNs). Given that we aim for a scalable solution, \ie when the image database would be the entire Web image corpus, we choose the AlexNet deep CNN architecture~\cite{KrSH12} as it offers a good trade-off between performance and speed.
% (\eg over the higher performing but far more computationally expensive Inception architecture~\cite{SLJ+14}).

We first get the $4k$-dimensional $fc7$ features, \ie the features after the second fully connected layer. In pursue for a more compact representation, we use dimensionality reduction to get the features to only $d=128$ dimensions. It has been shown that such embedding makes the image representation orders of magnitude more compact while minimally affecting search performance~\cite{BSCL14}. 

To reduce dimensionality we use PCA, learned using $fc7$ features from a public 100 Million YFCC100M image set~\cite{TSF+15}. As we plan on further compressing the visual features to just a few, using Locally Optimized Product Quantization~\cite{KaAv14}, we also permute the feature dimensions after PCA so that variance is balanced among all sub-vectors (see \S~\ref{subsec:scaling} for more details).

\vspace{.2cm}
\subsection{Selecting the Most Visually Similar Ad}
\label{subsec:selection}
% \label{subsec:select}
\vspace{.2cm}

% \head{Selecting the most visually similar ad.}
\noindent
Let $\cII = \{\It_1, \ldots, \It_n \}$ and $\cAA = \{ \At_1, \ldots, \At_m \}$ be the sets of visual features for $\cI$ and $\cA$ respectively, where $\It_k, \At_j \in \real^d$ and $k \in \{1,n\}, j \in \{1,m\}$. Getting the most visually similar ad requires finding the optimal ad image $\tilde{a} \in \cA$ so that 
\begin{equation}
\tilde{a} = \arg \min_{\At \in \cAA}  dist(a, \cII),
\label{eq:opt_a0}
\end{equation}
where $dist(a, \cII)$ is a function of dissimilarity between feature $a$ and \emph{feature set} $\cII$ of all images in the result set. There are multiple ways of defining such dissimilarity between a vector and a set. One would be to set $\tilde{a}$ to the ad that is the closest to one of the images, \ie
\begin{equation}
\tilde{a} = \arg \min_{\At \in \cAA}  \min_{\It \in \cII} || \At - \It ||,
\label{eq:opt_a1}
\end{equation}
where dissimilarity between two feature vectors is measured using the Euclidean distance. This way, the globally minimum dissimilarity between the ad feature set $\cAA$ and image feature set $\cII$ is kept. 

This is basically a nearest neighbor search in visual feature space, where we select the ad that is closest to \emph{any} of the images. Therefore, this formulation does not take into account the set as a whole and implies the danger of selecting an ad that matches an image which is not visually consistent with the rest of the set. 

Another, more robust, way of measuring this dissimilarity, would be by taking the whole set \emph{jointly} into account and selecting the ad that is closest to the set $\cII$ as a whole, or
\begin{equation}
\tilde{a} = \arg \min_{\At \in \cAA}  \sum_{\It \in \cII} || \At - \It ||.
\label{eq:opt_a2}
\end{equation}
We found that this is a better way of selecting visually congruent ads that blend better inside the result set. Therefore, unless otherwise stated we will use Equation (\ref{eq:opt_a2}) for selecting the best ad in the rest of this paper.

During ad selection, we sort all ads with proximity to the given image set and select the closest ad $\tilde{a}$. At no extra cost, we can keep the information on which is the most similar ad for each one of the images in set $\cI$. 

As we present in Section~\ref{subsec:place}, during ad placement, we also sort all images of set $\cI$ with proximity to the selected ad. These are dual problems, and we further require \emph{reciprocity} in nearest neighbor search for accepting the selected ad. In other words, we want the nearest neighbor of the selected ad in the image set $\cI$ to also have that specific ad as its nearest neighbor in set $\cA$ as  well.

%% file: tex/method_positioning.tex
\subsection{Positioning the Advertisement}
\label{subsec:place}
\vspace{.2cm}

\noindent
We investigated different advertisement positioning strategies, all of them assuming that we present results to the user in the typical way for image search, \ie on an 2d grid. 

Most of the presented strategies affect the ordering of images in different ways. For the case where the original ordering is considered very important, we also present a positioning strategy that does not affect that ordering. Figure~\ref{fig:placement} shows some of those strategies that are discussed below. Each white square is an image from set $\cI$, while the selected ad image is colored in dark cyan and its nearest images from set $\cI$ in lighter cyan.

Let the image indices in set $\cI$ also reflect the \emph{original ordering} of the images according to relevance to the textual query $q$. Below we present our different ad positioning strategies ordered in terms of how much they affect the initial ordering of set $\cI$. 

Let $\tilde{a}$ be the ad image selected as described in \S~\ref{subsec:selection}. Assuming that we want to maximize visual congruence by showing visually similar images nearby on the grid, it is important to also calculate the ordering of all images of set $\cI$ by visual proximity to the selected ad $\tilde{a}$. We therefore calculate set $\tilde{\cI} = \{ \tilde{\It_k} \}$, where $\forall k \in \{1,n-1\} : || \It_{i} - \tilde{a} || <  || \It_{i+1} - \tilde{a} ||$.

\input{fig/placement}

\subsubsection{Preserving the Original Image Ordering}
% \head{Preserving the original image ordering.} 
If preserving the original image ordering is important, the optimal way of placing the ad is right next to it most visually similar image $\tilde{\It_1}$. One can choose the side of placement (left or right) based on the dissimilarity of the neighboring images of $\tilde{\It_1}$ in the original ordering, and place it on the side of the neighbor with the largest similarity to $\tilde{a}$. 

\subsubsection{Altering the Ordering Locally Around the Ad} 
% \head{Altering the image ordering locally around the ad.} 
Allowing the original ordering to change, we get more freedom in placing the ad. One can place the ad image between its two nearest neighbors from set $\tilde{\cI}$, \ie between $\tilde{\It_1}$ and $\tilde{\It_2}$. This way we only change the position of $\tilde{\It_2}$ and keep the remaining images intact in terms of order. Figure~\ref{fig:placement}(a) presents a synthetic example of this strategy. 

Until now, we assume that images are ordered in a 1-dimensional
stream. More visually congruent positioning strategies can be achieved
if we also consider the 2d nature of the displayed image grid. We assume that the image grid is static and not responsive, \ie it does not change when the browser's window resizes. In the responsive case, one solution would be to re-compute orderings on the fly.

In this case the neighborhood of each cell can be defined in 4- or 8-way connectivity. Generalizing our previous placement, we can surround the ad image by its most similar images, either in 4 or all 8 directions.
Figure~\ref{fig:placement}(b) presents a synthetic example of this approach.

\subsubsection{Altering the Ordering Globally}
\emph{Stochastic Neighborhood Embedding} or \emph{t-SNE}~\cite{VaHi08}, is an embedding that aims in reducing the dimensionality of high-dimensional data to very few dimensions, \ie 2 or 3, while at the same time preserving distances in the original space. It was shown to be very successful for visualizing image datasets in a way that visual similarity is preserved. 

Such an embedding seems ideal for our case as well: Projecting all images and the ad image in that 2d space can provide us with data for placing not just the ad image among visually similar ones, but also every other image in the set. 

As shown in the synthetic example if Figure~\ref{fig:placement}(c), though, using this embedding does not constrain results on a grid and the output coordinates are arbitrary. 

A simple way of placing images based on their 2d coordinates is to \textit{greedily} select the closest grid position in 2d space for each image. That means, that each image will take the closest grid position if it is not already occupied. Priority is given to the advertisement image, to ensure its positioning among similar ones and images can be considered in ascending visual similarity from the ad.

The greedy approach may result in visually similar images being place far apart on the grid. To avoid that, simple density based clustering  can be efficiently run on the 2d image coordinates: We used a fast Mean Shift clustering~\cite{Cheng95}, which is an algorithm that automatically selects the final number of modes. Given image clusters, we can now greedily place images that correspond to the same cluster in adjacent positions of the grid. A synthetic example is shown in Figure~\ref{fig:placement}(d), where images of the same cluster are shown with the same color. 

The advertisement image is placed among images of the same cluster, thus ensuring visual coherency. If we see that the ad image is placed in a cluster just by itself, we may discard the ad, since this is an indication that it is not consistent enough visually with at least some of the images in the set. This way we get a sense of how visually close the ad image is, relevant to the visual consistency of the image set $\cI$ itself without the need for a parameter or fixed threshold. 

% An example of this type of ad rejection is shown in Figure~\ref{fig:overview}. If the selected ad was the one depicted with gray fill in the second box from the right, it would be rejected; it corresponds to the yellow cluster that only contains one image.

Although slightly computationally expensive, the t-SNE embedding and
clustering-based algorithm can run in just milliseconds if we use fast
versions of the algorithms~\cite{VanD14,VeSo08b} and limit their input
to the first page of image search results, \ie a few dozen
images. This is not restricting in practice as in any case we would
want to place our ad image within the first page of results. For our
study, we chose the greedy placement to make the overall look more
consistent with the other conditions (see
\S~\ref{sec:experiments}). We do not evaluate the different
placements in the current experiments, 
leaving this for future work.
%%% Local Variables:
%%% mode: latex
%%% TeX-master: "../www16"
%%% End:

%% file: fig/placement.tex
\begin{figure*}
\centering
\begin{tabular}{cccc}

% ------------------------------------------------------------
% ------------------------------------------------------------
% a
% ------------------------------------------------------------
\begin{tikzpicture}[scale=.5]
\foreach \x in {0,...,6}
	\foreach \y in {0,...,4}
		\draw (\x,\y) rectangle (\x+1,\y + 1);
\filldraw[cyan!51!white, draw=black] (5,2) rectangle (6,3);
\filldraw[cyan!31!white, draw=black] (3,2) rectangle (4,3);
\filldraw[cyan, draw=red] (4,2) rectangle (5,3);
              \draw[black,thick] (4,2) -- (5,3);
              \draw[black,thick] (5,2) -- (4,3);
\end{tikzpicture} 
% ------------------------------------------------------------
% ------------------------------------------------------------
&
% ------------------------------------------------------------
% ------------------------------------------------------------
% b
% ------------------------------------------------------------
\begin{tikzpicture}[scale=.5]
\foreach \x in {0,...,6}
	\foreach \y in {0,...,4}
		\draw (\x,\y) rectangle (\x+1,\y + 1);
\foreach \x in {3,...,5}
	\foreach \y in {1,...,3}
		\filldraw[cyan!31!white, draw=black] (\x,\y) rectangle (\x+1,\y + 1);
\filldraw[cyan!61!white, draw=black] (5,2) rectangle (6,3);
\filldraw[cyan!61!white, draw=black] (3,2) rectangle (4,3);
\filldraw[cyan!61!white, draw=black] (4,3) rectangle (5,4);
\filldraw[cyan!61!white, draw=black] (4,1) rectangle (5,2);
\filldraw[cyan, draw=red] (4,2) rectangle (5,3);
              \draw[black,thick] (4,2) -- (5,3);
              \draw[black,thick] (5,2) -- (4,3);
\end{tikzpicture} 
% ------------------------------------------------------------
% ------------------------------------------------------------
&
% ------------------------------------------------------------
% ------------------------------------------------------------
% c
% ------------------------------------------------------------
\begin{tikzpicture}[scale=.5]
\foreach \x in {0,...,6}
	\foreach \y in {0,...,4}
		\draw[white] (\x,\y) rectangle (\x+1,\y + 1);

% ad cluster
\filldraw[cyan!40!white, draw=black] (2.3,3) rectangle (2.8,3.5);
\filldraw[cyan!40!white, draw=black] (1.4,3) rectangle (1.9,3.5);
\filldraw[cyan!40!white, draw=black] (1.3,4) rectangle (1.8,4.5);
\filldraw[cyan!40!white, draw=black] (1.5,4.3) rectangle (2,4.8);
\filldraw[cyan!40!white, draw=black] (1.9,4.5) rectangle (2.4,5);
\filldraw[cyan!40!white, draw=black] (2.7,4.4) rectangle (3.2,4.9);
\filldraw[cyan!40!white, draw=black] (2.5,3.8) rectangle (3,4.3);
\filldraw[cyan!40!white, draw=black] (1.8,3.8) rectangle (2.3,4.3);
\filldraw[cyan, draw=red] (2.1,4.2) rectangle (2.6,4.7);
				\draw[black,thick] (2.1,4.2) -- (2.6,4.7);
              	\draw[black,thick] (2.1,4.7) -- (2.6,4.2);
\filldraw[magenta!40!white, draw=black] (4.1,4.1) rectangle (4.6,4.6);
\filldraw[magenta!40!white, draw=black] (4.5,2.8) rectangle (5,3.3);
\filldraw[magenta!40!white, draw=black] (4,3.4) rectangle (4.5,3.9);
\filldraw[magenta!40!white, draw=black] (4.3,4.4) rectangle (4.8,4.9);
\filldraw[magenta!40!white, draw=black] (5.1,3.9) rectangle (5.6,4.4);
\filldraw[magenta!40!white, draw=black] (5.2,3.5) rectangle (5.7,4);

\filldraw[yellow!60!white, draw=black] (5.5,2.4) rectangle (6,2.9);

\filldraw[red!50!white, draw=black] (5.0,0.8) rectangle (5.5,1.3);
\filldraw[red!50!white, draw=black] (4.7,1.2) rectangle (5.2,1.7);

\filldraw[green!40!white, draw=black] (1.6,1.2) rectangle (2.1,1.7);
\filldraw[green!40!white, draw=black] (2.2,0.7) rectangle (2.7,1.2);
\filldraw[green!40!white, draw=black] (1.9,1.4) rectangle (2.4,1.9);
\filldraw[green!40!white, draw=black] (2.5,1.2) rectangle (3,1.7);
\filldraw[green!40!white, draw=black] (1.5,0.2) rectangle (2,0.7);

\end{tikzpicture} 
% ------------------------------------------------------------
% ------------------------------------------------------------
&
% ------------------------------------------------------------
% d
% ------------------------------------------------------------
\begin{tikzpicture}[scale=.5]
\foreach \x in {0,...,6}
	\foreach \y in {0,...,4}
		\draw (\x,\y) rectangle (\x+1,\y + 1);
        
\foreach \x in {0,...,4}
	\foreach \y in {2,...,4}
		\filldraw[cyan!40!white, draw=black](\x,\y) rectangle (\x+1,\y + 1);
\filldraw[cyan, draw=red] (1,3) rectangle (2,4);
              \draw[black,thick] (1,3) -- (2,4);
              \draw[black,thick] (1,4) -- (2,3);
\foreach \x in {5,...,6}
	\foreach \y in {1,...,4}
		\filldraw[magenta!40!white, draw=black](\x,\y) rectangle (\x+1,\y + 1);
\filldraw[magenta!40!white, draw=black](4,2) rectangle (5,3);
\filldraw[magenta!40!white, draw=black](4,1) rectangle (5,2);
\foreach \x in {0,...,3}
	\foreach \y in {0,...,1}
		\filldraw[green!40!white, draw=black](\x,\y) rectangle (\x+1,\y + 1);
\filldraw[red!50!white, draw=black](4,0) rectangle (5,1);
\filldraw[red!50!white, draw=black](5,0) rectangle (6,1);
\filldraw[yellow!60!white, draw=black](6,0) rectangle (7,1);

\end{tikzpicture} \\
% ------------------------------------------------------------
% ------------------------------------------------------------

(a) & (b) &(c) & (d)
\end{tabular}

\caption{Different positioning strategies. The selected ad image is represented as a dark cyan colored square with a red border and a cross; all other squares correspond to images from set $\cI$. White squares are images placed in the original ordering. (a): Positioning with minimal change of the original ordering; the ad is placed right next to its nearest images from set $\cI$ and at most one image from $\cI$  would change its place. (b): Altering the image order locally around the ad; the images in lighter cyan would be repositioned around the ad in ascending order of $k$ as in set $\tilde{i_k}$. (c): Density based clustering in 2d space after t-SNE projection. Different colors correspond to different clusters. (d): Placing image clusters on the grid.}
\label{fig:placement}
\end{figure*}
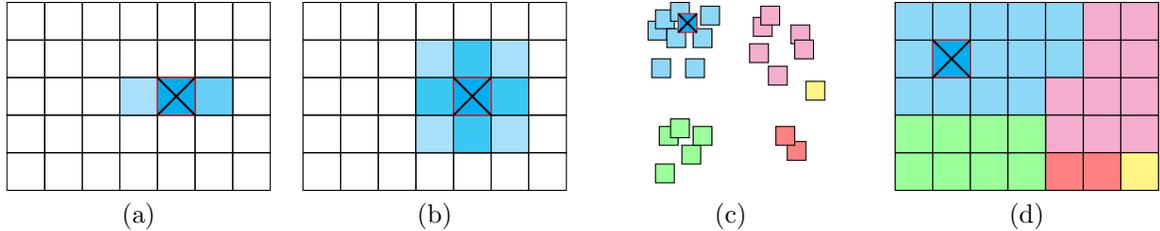

%% file: tex/method_scaling.tex
\subsection{Scaling by Compressed Image Signatures}
\label{subsec:scaling}
\vspace{.2cm}

\noindent
We want our approach to be applicable in real-time for web-scale image search. In the typical scenario, images are indexed by their surrounding textual metadata and, given a query, are ordered with relevance to those metadata through text matching~\cite{BYRN11}. We therefore only need to deal with the top-$k$ images returned by the index. 

However, visual feature extraction is a very computationally heavy process that cannot run on the fly even for a few dozen photos. That means that we need to have visual features stored and available for the whole image dataset, which in a web-scale image search scenario would contain billions of images crawled from the web. The CNN features we use are powerful; however, they require storing a vector of 128 floating point numbers for each image. Although for most content-based applications this is considered a pretty compact signature, when scaling to $10^{10}$ photos, every byte counts. 

Recent quantization approaches for large scale nearest neighbor search can provide very compact signatures with only a small drop in search performance~\cite{JeDS11, GHKS14, KaAv14}. Such methods perform much better than hashing, and, unlike hashing, using allow for a \emph{reconstruction} of the original vector from the compressed domain, giving a much finer way of measuring (dis-)similarity. We therefore choose to use a recently proposed extension of Product Quantization~\cite{JeDS11} for compressing visual features named \emph{Locally Optimized Product Quantization} or \emph{LOPQ}~\cite{KaAv14}. Keeping in mind all the different possible use cases for this content-based image signature outside this specific application of ad placement\footnote{The same signature can have multiple uses, \eg deduplication, diversification, visual similarity search, \etc.}, we specifically choose to use the Multi-LOPQ approach of~\cite{KaAv14} that also enables efficient indexing and search.

Each vector is split into $M$ sub-vectors and is quantized independently using one Byte per subvector. The codebook $\cC$ is considered to be composed of a Cartesian product of sub-codebooks, one for each of the sub-spaces. In the LOPQ extension, a two-level quantization is used; where codes and sub-spaces are optimally computed locally in the second quantization level in a way that variance is balanced among sub-spaces (we refer the reader to~\cite{KaAv14} for more details). 

In the end, we represent each feature vector $x \in \real^d$ with a code $c = \{c_1, \ldots, c_M\}$ of size $M$, where  $c_i$ is a byte-size representation for each sub-vector, typically the index of the closest sub-codebook centroid for the $i$-th sub-vector
% . Apart from the M-Byte code, we also need to store the two coarse quantization indices from the first, coarse quantization step, that typically require $3-4$ extra Bytes.
, together with the two coarse quantization indices from the first, coarse quantization step.

Advertisement images are even in the web-scale scenario in the order of thousands. As we further restrict ourselves to \emph{relevant} advertisements, the average number of ads to consider is even lower. In such a scenario compressing their signature is not really worth the further approximation induced by it. Of course, for the case where more than a couple thousand ads are often being considered for placement, one can also quantize the ad signatures and use the Multi-LOPQ approach for search~\cite{KaAv14}. We therefore choose not to quantize advertisement images and use \emph{asymmetric distance computations} for approximate nearest neighbor search, \ie the \textit{ADC} version of~\cite{JeDS11}.

Given a code and all quantizers, one can project every compressed feature back to the original $d$-dimensional space. After projecting, advertisement selection and placement proceed as in \S~\ref{sec:method}.
In the next section we evaluate the numerical errors due to quantization, by measuring how many times the same advertisement image is picked as closest given an image set, with and without compression. We proceeded using the compressed signatures, which proved to be effective in selecting ad images that are perceptually consistent with the set.

%% file: tex/experiments.tex
\section{Experiments}
\label{sec:experiments}
\vspace{.2cm}

% --------------------------------------------------------------------------
\begin{figure}[h!]
	\begin{center}
	\begin{subfigure}[t]{\linewidth}
		\centering\includegraphics[width=\linewidth]{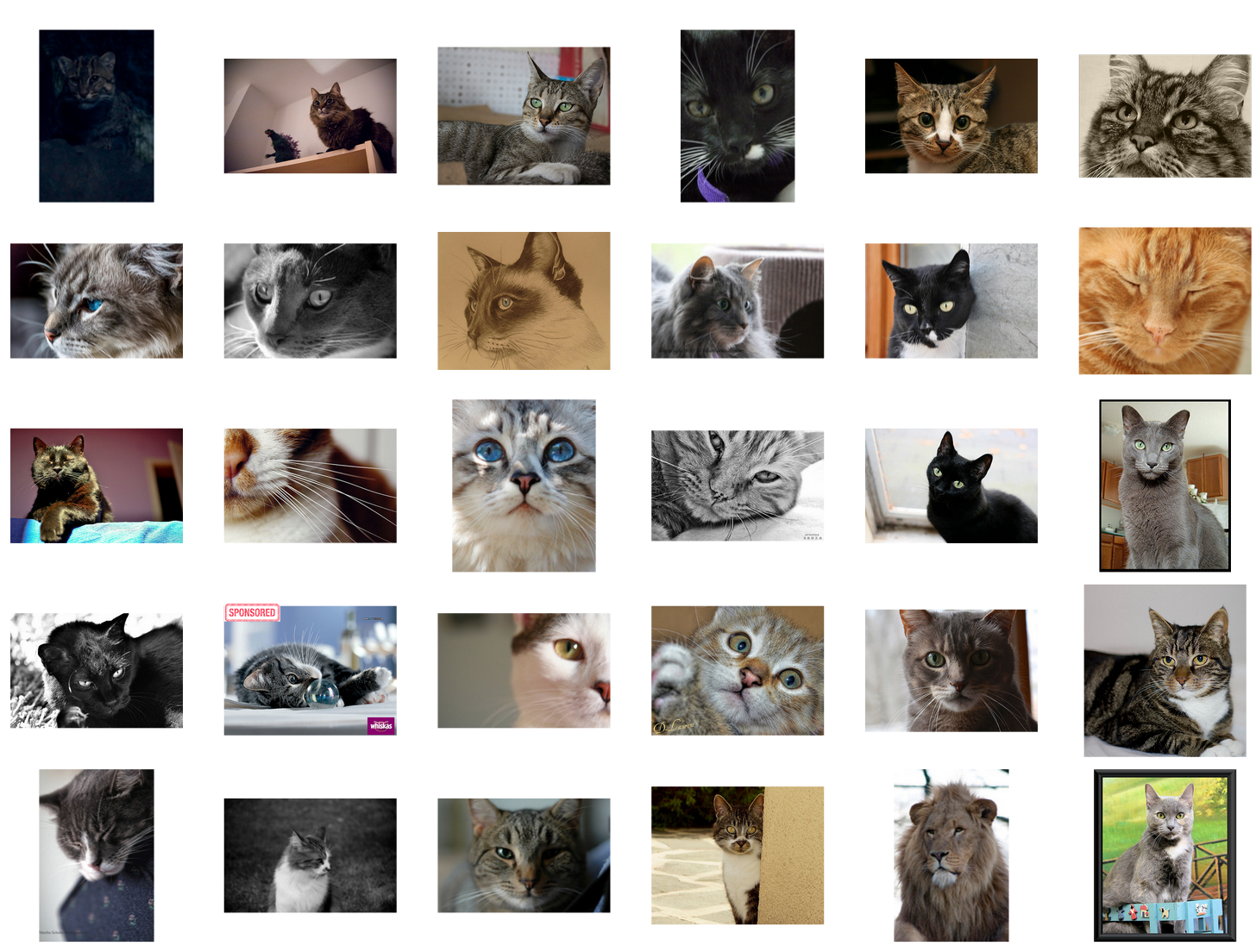}
% 		\centering\includegraphics[width=\linewidth]{fig/fashion_fashion_optimal}
% 		\vspace{-0.5cm} 
        \caption{\textit{Most similar} ad selection, optimal positioning}
		\label{fig:clust}
	\end{subfigure}%

	\vspace{.3cm}

	\begin{subfigure}[t]{\linewidth}
		\centering\includegraphics[width=\linewidth]{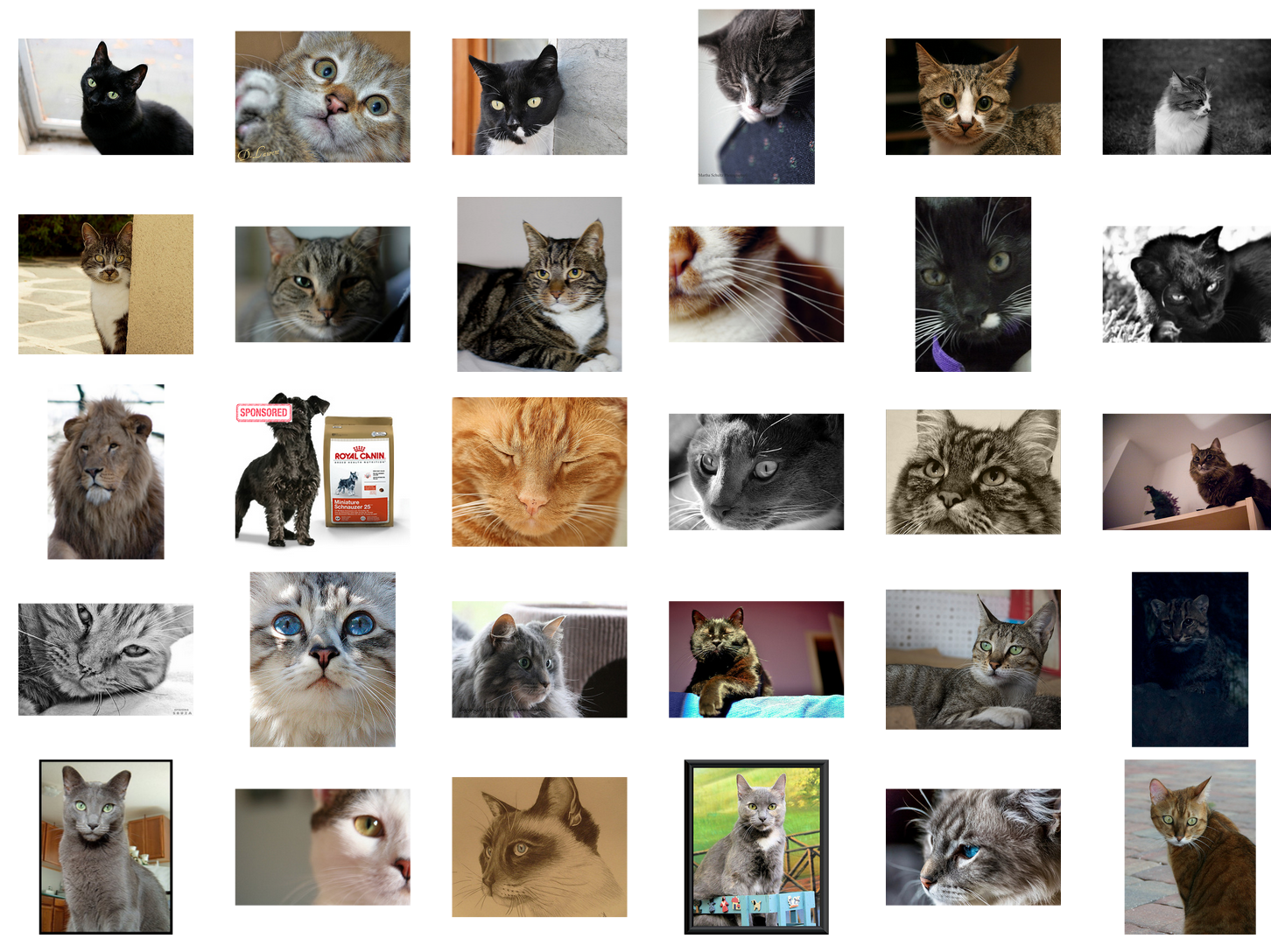}
% 		\centering\includegraphics[width=\linewidth]{fig/fashion_fashion_random}        
% 		\vspace{-0.5cm}
        \caption{Random ad selection and positioning}
		\label{fig:recsoc}
	\end{subfigure}%

	\vspace{.3cm}

	\begin{subfigure}[t]{\linewidth}
		\centering\includegraphics[width=\linewidth]{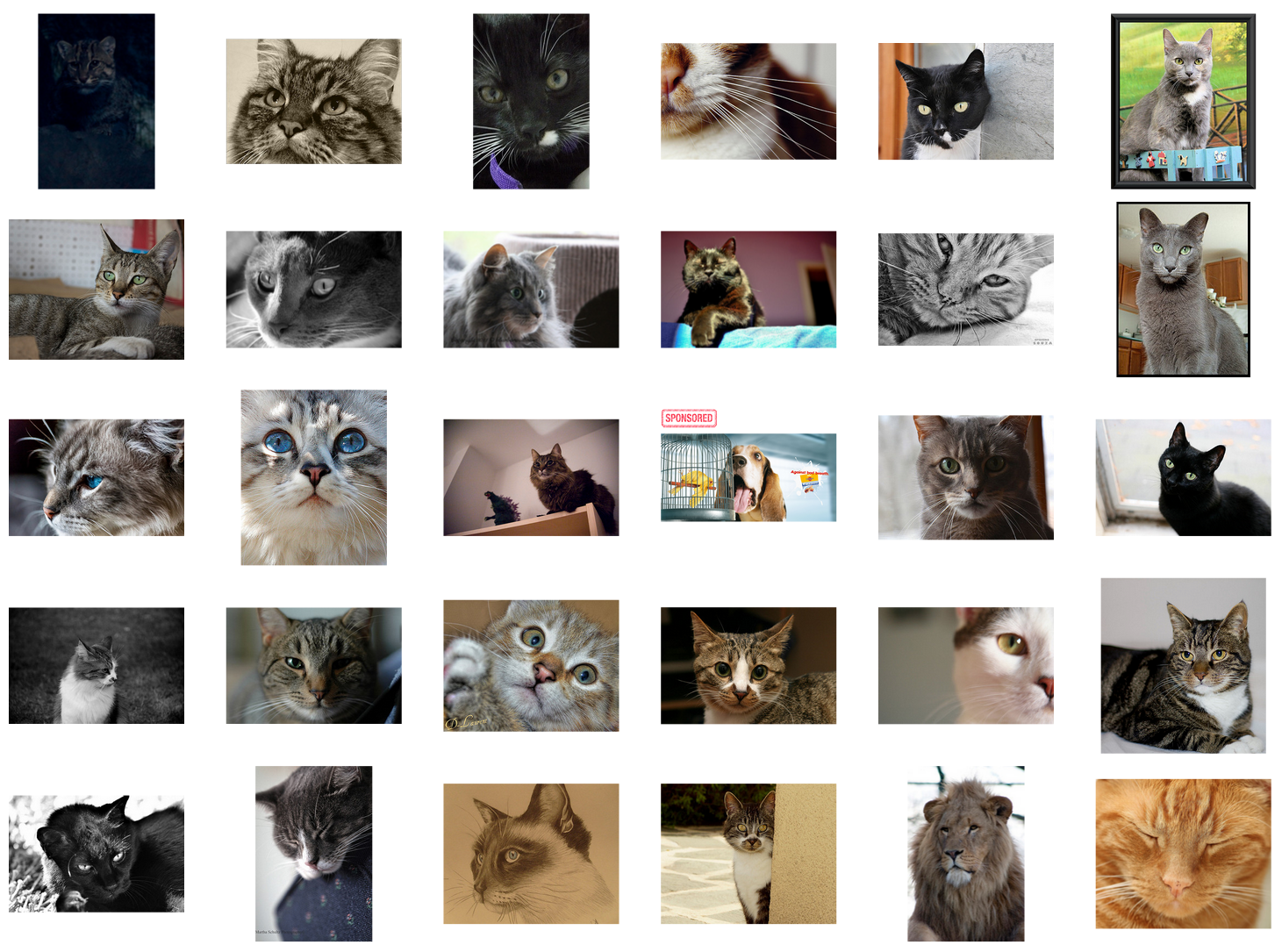}
% 		\centering\includegraphics[width=\linewidth]{fig/fashion_fashion_worst}
% 		\vspace{-0.5cm}
        \caption{Least similar ad selection and positioning}
		\label{fig:recvis}
	\end{subfigure}%

	\caption{The three different conditions presented to workers for query ``cats''. The ad is marked with the word \textbf{sponsored}.\textit{ Best viewed magnified on a monitor}. }
	\label{fig:example}
	\end{center}
\end{figure}
% -------------------------------------------------------------------------- 

% experiments intro
Now we present the experimental evaluation of the proposed ad placement approach. Before explicitly measuring the effectiveness of our optimal ad placement in a live field experiment, we choose to run a large-scale study on Amazon Mechanical Turk (AMT). 

We devised a hard scenario to validate our approach: Given an image search result set and a set of ads that are \emph{all relevant} to the image set \emph{and} to the users interests, we tested three different ad placements or \textit{conditions}, one using the proposed approach, a random one, and one where we tried to make the ad to stand out of the result set as much as possible.

As mentioned before, in order to simulate targeted advertising, in all three conditions the ad was relevant to the images and the user, \eg we place a car-related ad in an image search result set for query ``best car 2015'' and requiring people to be interested in cars in order to answer our survey. 

% In Section~\ref{subsec:exp_prelim} we present the evaluation protocol, the dataset used and some preliminary experiments that validate and quantify the approximation of nearest neighbor search using the compressed image features. We present our analysis and most significant findings from the survey in Section~\ref{subsec:exp_amt}. 

% evaluation protocol, ad dataset stats and preliminary experiments
\input{tex/experiments_preliminary}

% main AMT study
\input{tex/experiments_amt}

%% file: tex/experiments_preliminary.tex
\subsection{Evaluation Protocol and Validation}
\label{subsec:exp_prelim}

% --------------------------------------------------------------------------
% dataset
To test our hypothesis, we curated an advertisement dataset. We selected image-based ads of high aesthetic quality from five generic topics of interest: \textit{animals}, \textit{cars}, \textit{fashion}, \textit{movies} and \textit{TV Series}. We gathered 150 ads in total; the number of ads per topic are presented in the top row of Table~\ref{tab:ad_dataset}. 
% --------------------------------------------------------------------------

% --------------------------------------------------------------------------
% Ad dataset table
\begin{table}
\centering
\begin{tabular}{r ccccc}
%% \toprule
& \multicolumn{5}{c}{\textbf{Topic}} \\
\cmidrule(r){2-6}
& \textit{Animals} & \textit{Cars} & \textit{Fashion} & \textit{Movies} & \textit{TV}  \\ 
\midrule
Ads & 23 & 48 & 45 & 16 & 18 \\ % & Total 150 \\ 
% \midrule
Acc. & 0.74 & 0.69 & 0.67 & 0.81 & 0.81  \\ % & Total 0.75 \\ 
\bottomrule
\end{tabular}
\caption{Top row: Breakdown of our ads dataset showing number of ad images per topic. The total number of ads in the dataset is 150. Bottom row: Average nearest neighbor accuracy per topic using the compressed versions of the features versus the original (uncompressed) ones. The average for the whole dataset is $0.75$.}
\label{tab:ad_dataset}
\end{table}
% --------------------------------------------------------------------------

% --------------------------------------------------------------------------
% image search using flickr
To simulate image search, we query the public Flickr API\footnote{\url{https://www.flickr.com/services/api/}} for photos with different search terms related to the 5 topics of interest. We specifically used the \texttt{photos.search} method ordered by relevance. 
% --------------------------------------------------------------------------

% --------------------------------------------------------------------------
% visual representation protocol
To represent the visual content of ads and image search results, we follow the process presented in \S~\ref{subsec:selection}. We extract compact CNN features, reduce their dimensionality down to 128 dimensions and compress them using a Multi-LOPQ~\cite{KaAv14} model with $M=16$ and coarse vocabularies of size $2^{13}$. This means that we need to store in total just $128 + 26 = 154$ bits per image, \ie approximately 20 Bytes. We also L2 normalize the reduced features before compression and nearest neighbor search. 
% --------------------------------------------------------------------------

% --------------------------------------------------------------------------
% statistical analysis tools used
As the most interesting survey questions we asked had Likert scale answers, \ie correspond to ordinal data, we cannot rely on classic continuous-space statistical estimators like mean and standard deviation~\cite{HoWC13}. We therefore report the Mann-Whitney U test results for significance estimation, as well as percentages per condition.

For the time statistics in Table~\ref{tab:stats} we used robust estimates for mean and standard deviation~\cite{Huber64} to exclude outliers.

% --------------------------------------------------------------------------
% NN search approximation with compression
To quantify the approximations induced by compression in nearest neighbor search, we also present some quantitative results on our ad dataset in the last row of Table~\ref{tab:stats}. As argued in \S~\ref{subsec:scaling}, we compress the results image set features only and not the ad features. 

For evaluation, we run 100 generic queries in the Flickr API and report the average accuracy for ad selection for each of the 5 topics, \ie how many times the ad selected as most similar was the same when using the uncompressed features and the compressed features. We see that 3 out of four times the same ad was picked. 

These results are purely quantitative and just confirm the results reported in~\cite{KaAv14}, \ie  computing visual similarity in the compressed domain approximates similarity in the original domain pretty well. 

However, as no groundtruth can exist to tell us which ad \emph{should} be picked, approximation results cannot give us any guaranties regarding the selected ad's distractingness factor or relevance to the set. To measure such aspects, we run a large crowd sourced study in AMT.
In this study we used the compressed image features, as this is the only viable option for a real-time and web-scale scenario. We used psiTurk\footnote{\url{https://psiturk.org/}} to run the study.

% More details on the study as well as analysis of the results can be found in Section~\ref{subsec:exp_amt}. 

% --------------------------------------------------------------------------

%% file: tex/experiments_amt.tex
\subsection{Large Scale Study in AMT}
\label{subsec:exp_amt}

We gathered result sets for 8 queries that are semantically relevant to the 5 categories of our ads dataset. For each of the 8 sets we created three conditions. Each worker was presented with \textit{only one condition from one set}, selected randomly. We did this as we did not want to let the workers know that we were testing ad placement and be biased.

The three conditions we experimented on were:
\begin{description}
\item[Most Similar] Optimal selection and positioning of the most similar ad using our method.
\item[Random] Random selection and positioning.
\item[Least Similar] Selection of the least similar ad and positioning it among the most dissimilar images.
\end{description}

To simulate targeted advertising and assume that the ad matches the user's interests, we further required the workers to be interested in the ads topic, \eg we titled the survey that presented fashion-related ads as ``A survey for people who are into fashion''. 

We used the greedy positioning after 2D projection for the \textit{Most Similar} and \textit{Least Similar} conditions, where in the latter case we surround the ad with its most distant images from the set \wrt visual similarity.

We set the following 3 requirements for AMT workers: To have more than $95\%$ of their HITs accepted, to have completed more than $1,000$ HITs and, to be located in the US\@. Based on them we accepted the HITs from 896 workers.
This implies that we showed each condition of each query (\ie in total we had 24 different versions) to about 37 workers. Of those who chose to declare their gender, $43\%$ were female and $57\%$ male. In terms of age distribution, $22\%$ of the workers were under 26 years old, $49\%$ were between 26-35 years old, $16\%$ were between 36-45 years old and $12\%$ were over 45 years old.

% --------------------------------------------------------------------------
% survey stats table
\begin{table*}[t]
\centering
\def\arraystretch{1.2}
\begin{tabular}{l ccc} 
   & \textbf{Most similar} & \textbf{Random} & \textbf{Least similar} \\ \hline
Workers per case & 302 & 288 & 306  \\
\midrule
Saw an ad & 39.7\% & 38.5\% & 38.6\%  \\
Saw the correct ad & 86.7\% & 63.1\% & 79.7\%  \\ 
\midrule
Mean time spent & $54.4$s & $54.1$s & $52.2$s  \\
% Time spent at image set page (sec) & $54.432$ \tiny{($\pm 24.802$)} &$54.160$ \tiny{($\pm 24.198$)} &$52.223$ \tiny{($\pm 25.618$)} \\
\bottomrule
\end{tabular}
\caption{Statistics from the AMT study.  Time spent carried had a standard deviation of $\pm 24.8$, $\pm 24.2$, $\pm 25.6$ respectively across the Most, Random, and Least Similar categories.}
\label{tab:stats}
\end{table*}
% --------------------------------------------------------------------------

\subsubsection{Study Protocol}

The study consisted of two pages: the \textit{image set page} and the \textit{questionnaire page}.
On the image set page, the worker would see the result set together with two questions that would ``force'' them to actually pay attention to the images in the set. The first question was different for each result set, and would contain something in reference to the content of the images shown, usually in the format ``How many images depict \textit{X}'', where \textit{X} would be \eg ``brown dogs'', ``red trucks'', ``dresses''.

For example, Figure~\ref{fig:example} shows the three conditions for the query ``fashion'' from the homonymous topic. For all three conditions shown, the first question was ``How many images depict dresses?''.

The second and final question of the first page was shared among all queries and was the question ``What is an appropriate title for this image set?''. After answering both questions the workers would proceed to the questionnaire page, \textit{without the option of seeing the image set again}.
Up to the questionnaire, nothing would reveal to the user the actual purpose of the survey.
The questionnaire consisted of 11 questions in total; the complete list can be seen in Table~\ref{tab:questions}.

Some of those had open-ended text fields for answers, others were multiple choice and others in the 5-choice Likert scale: \textit{Strongly Disagree, Disagree, Neutral, Agree}, and \textit{Strongly Agree}. Below, we analyze the most important of the Likert scale questions independently.
Of critical importance was question (Q2) ``I recall seeing a sponsored ad from:'' that had five multiple-choice answers: The first three choices were query-dependent and were known brands from the specific topic, one of which also corresponded to the brand of the ad that the worker saw. The forth choice was ``other brand'' and the fifth was ``I didn't see an ad''.

% --------------------------------------------------------------------------
% table with all questions of the survey
\begin{table}
\centering
\resizebox{\linewidth}{!}{%
\begin{tabular}{rl}
\toprule
\multicolumn{2}{l}{\hspace{.3cm}\textbf{Image set page}} \\
\textbf{I1} &  \textit{Image set specific question}  (Text)  \\
\textbf{I2} &  What is an appropriate title for this image set?  (Text)  \\
\multicolumn{2}{l}{\hspace{.3cm}\textbf{Questionnaire page}} \\
\textbf{Q1} &  Did anything seem out of place? If yes, what?  (Text) \\
\textbf{Q2} &  I recall seeing a sponsored ad from  (Multiple Choice) \\
\textbf{Q3} &  Can you recall what was being advertised?  (Text) \\
\textbf{Q4} &  The sponsored ad was relevant to my interests  (Likert) \\
\textbf{Q5} &  The sponsored ad was relevant to the image set  (Likert) \\
\textbf{Q6} &  The sponsored ad was clearly marked  (Likert) \\
\textbf{Q7} &  The sponsored ad was distracting and out of place  (Likert) \\
\textbf{Q8} &  The ad experience was of very high quality  (Likert) \\
\textbf{Q9} &  I am very familiar with the advertised brand  (Likert) \\
\textbf{Q10} &  Gender information  (Multiple Choice) \\
\textbf{Q11} &  Age information  (Multiple Choice) \\
\bottomrule
\end{tabular}
}
\caption{List of questions asked in our large scale Amazon Mechanical Turk study. We show the answer type in parenthesis after each question.}
\label{tab:questions}
\end{table}
% --------------------------------------------------------------------------

\subsubsection{Analysis of Results}

Some basic statistics are shown in Table~\ref{tab:stats}. In the first row, we show the number of workers per condition, which is fairly balanced for all three. The second row reports the percentage of workers that answered positively in (Q2), \ie chose one of the first four choices. It is notable that in all three conditions, about $39\%$ of the workers claimed that they saw the ad. It is really essential for this number to be constant, as this shows that \textit{showing an ad that is visually similar to the image set does not harm the recognition of the ad}. 
% As we show below, it actually affects the experience in a positive way.
The third row presents the percentage of workers that answered (Q2) \textit{correctly} per condition, \ie responded that they saw the brand that was indeed been advertised in the set they saw. What is notable here is that the percentage of workers that could identify the brand correctly was \textit{highest in the condition where the ad was selected by the proposed algorithm}. The fact that the percentage is also high in the \textit{Least Similar} condition confirms related work that ad obtrusiveness correlates with ad recognition~\cite{CoSF12}.  The final row reports the mean and standard deviation\footnote{To exclude outliers (\eg we saw that the slowest worker spent over 25 minutes at the image set page) we used robust estimates~\cite{Huber64} for the mean and standard deviation reported.} of the time spent on the image set page for the workers that claimed to have seen an ad. We see that the time is practically constant in all three cases, while the median for all conditions is 51.8 seconds. 
% The median for workers that claim to have \textit{not} seen an ad is \alert{(AymanS? A number here, pleeeeeeease???)}. The median for time spent in the questionaire page was 63.5 (\alert{(AymanS? Another number here?  - See what happens when you do the analysis yourself?? :) )} ???) for workers who saw (didn't see) an advertisement. 

\begin{figure*}[t]
\begin{center}
  \includegraphics[width=\linewidth]{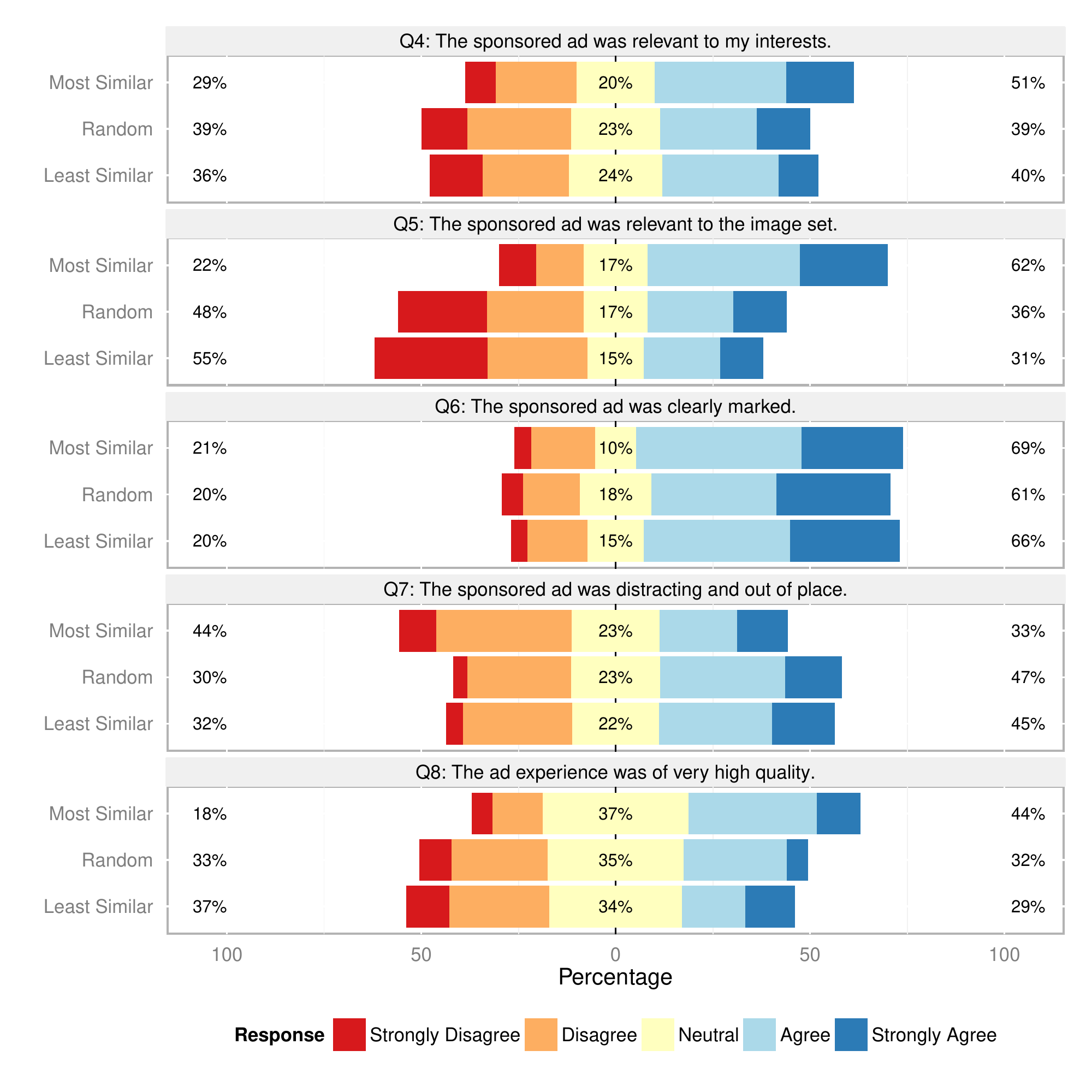}  
\end{center}
\vspace*{-.5cm}
\caption{Stacked percentages per condition for five of the questions
  asked in our survey. Each subplot presents percentages for the
  question shown in its caption. We see that the proposed approach,
  \ie the condition denoted \textit{Most Similar}, gives favorable
  results, while not affecting the advertisement's recognition
  negatively (Q6).}
\label{fig:plots}
\end{figure*}

In Figure~\ref{fig:plots} we show the percentages of each of the five possible responses per condition for the five most important questions/statements. All five are in Likert scale, so, to reject the null hypothesis~\cite{HoWC13} we also report p-values using the Mann-Whitney U test, when considering the three conditions in pairs. 
We always report p-values for the three pairs $p_1 = $ (Most Similar, Random), $p_2 = $ (Most Similar, Least Similar) and $p_3 =$ (Random, Least Similar) in this order.
Below, we analyze the most interesting findings for some of the questions/statements independently. 

\head{(Q8) The ad experience was of very high quality}. The percentages for this question presented in Figure~\ref{fig:plots} show significant signs that our basic hypothesis was right. Using visual similarity to pick the ad greatly affects the ad experience of the workers. 

In fact, $44.5\%$ of the workers that were shown the \textit{Most Similar} condition chose Agree or Strongly Agree for the statement, while the same percentages for the \textit{Random} and \textit{Least Similar} conditions are $31.5\%$ and $28.8\%$ respectively. Percentages are reverse for Disagree and Strongly Disagree; only $18.4\%$ of the workers shown the condition that was selected by our algorithm found their ad experience of low quality, while the percentages are $33.3\%$ and $37.2\%$ for the other two cases, respectively.

The Mann-Whitney test confirms the statistical significance of the results. It gives us strong evidence to reject the null hypothesis as the p-values are $0.011, 0.006$ and $0.747$ for the three pairs respectively, where the first two pairs ($p_1$ and $p_2$) compare the proposed approach to the other two conditions. Picking the \textit{Least similar} ad was in fact not so much different than picking at \textit{Random} in terms of user perception of ad experience.

\head{(Q7) The sponsored ad was distracting and out of place}. The Mann-Whitney test also provides us with some weak evidence to reject the null hypothesis in the distractingness question (the $p$-values for the three pairs are $0.0565$, $0.0531$ and $0.9613$ respectively). This confirms the positive skew towards the \textit{Most Similar} condition as seen in the percentages for this question in Figure~\ref{fig:plots}. About $44.0\%$ of the workers choose to Disagree or Strongly Disagree with the statement, while the same percentages are $30.6\%$ and $32.2\%$ for the other two conditions. 

\head{(Q5) The sponsored ad was relevant to the image set}. 
The results for this statement were also significant and interesting. It seems that most workers ($62.1\%$) deem the ad relevant to the image set when it was visually consistent with its surroundings, but not for the other two conditions where there is no visual congruence. Workers disagreed on the statement $47.7\%$ and $55.0\%$ for the other two conditions, respectively.

This result is especially interesting, because \emph{all ads} were
chosen to be relevant or targeted to the image set in a topical sense. With the general notion of ad relevance in text-based
ads, all of the ad dataset ads could have been considered
``targeted''.  It seems that people perceive relevance differently
when it comes to images, and it comes natural to them to
  interpret relevance in a visual way than in a topical way.

  % This is in line with previously published studies on the
  % subject~\alert{AymanF??? You mentioned a citation confirming
  % this.}.

  The $p$-values show the strong statistical significance for the
  \textit{Most Similar} condition versus the other two: they are
  $5 \cdot 10^{-5}, 4 \cdot 10^{-7}$ and $0.328$ for the three pairs,
  respectively. As we mention in future work, it would be very
  interesting to see what would happen if we take semantic relevance
  out of the equation and only take into account visual relevance for
  ad placement. We leave this experiment for future work.

  \head{(Q4) The sponsored ad was relevant to my interests}.  As
  mentioned in the survey protocol, we required workers to be
  interested in the general topic of the ads that we would show
  them. This question was therefore more a sanity check than anything
  else. Percentages are in all three conditions in favor of agreement
  with the sentence. Still, positive agreement with the sentence is
  visibly higher for the \textit{Most Similar} one over the other
  two. There is only weak evidence to reject the null hypothesis,
  however (the p-values for pairs $p1$ and $p2$ are $0.062$ and
  $0.074$ respectively).

\head{(Q6) The sponsored ad was clearly marked}.
As we saw from the stats of Table~\ref{tab:stats}, ad recognition is not affected by the visual congruence of the ad. We had marked the ad images with the word ``Sponsored'' in the experiment, and as we see from the percentages shown for this question in Figure~\ref{fig:plots}, most workers seem to agree with this sentence. Note that visual congruence once again does not affect how the advertisement is perceived.

% --------------------------------------------------------------------------
% Likert plots
% ... are in conclussions.tex, otherwise the would go after the References  (Ric: moving them to here so they appear earlier)
%

% --------------------------------------------------------------------------
% Likert plots

%%% Local Variables:
%%% mode: latex
%%% TeX-master: "../www16"
%%% End:

%% file: tex/discussion.tex
\section{Discussion}
\label{sec:discussion}
\vspace{.2cm}

\noindent
Here we discuss related issues as well as novelty aspects of our approach. Extensions mentioned below, like multiple advertisements, diversification of results or selection that takes into account other prior weights/modalities are not investigated in the current paper; some are left out of the evaluation for clarity and consistency, and others for future work. 
However, we think it is important to further discuss some aspects or implications of our approach as well as clarify novelty aspects.

\head{Multiple advertisements}. 
All the aforementioned approaches can be extended to selecting and placing multiple advertisements. For example instead of selecting the minimum distance ad from Equation (\ref{eq:opt_a2}), one can get the $k$-dissimilar ones by sorting the set of dissimilarities $\{ \sum_{\It \in \cII} || \At_j - \It || \}, \At_j \in \cAA$ and keeping the top-$k$ indices. This may complicate some of the naive placement strategies, but the t-SNE embedding based ones can easily cope with multiple ad placements.

\head{Considering multiple modalities}. The original ordering of the
images is essentially a multimodal ordering\footnote{Image search
  ordering does not traditionally take visual appearance into account
  explicitly, although such techniques may be used for deduplication.}
of the images based on text, metadata, page-rank, and so on.  In that
sense, the visual appearance of the ad can be considered as another
modality and incorporated in the joint ordering, together with image \textit{aesthetics}, a modality that we chose not to investigate.
Similarly, prior weights for advertisement image preference (\eg based on financial priors) can be incorporated in the selection and placement process. In such cases, though, one would need to take extra measures for ensuring that visual congruence is not heavily affected. 

\head{Financial implications}. 
% \alert{AymanF.? If you can sell this part nice, we maybe can avoid submitting in the search track that last year had 8.5\% acceptance rate...}
The economic and financial implication of monetizing image search while preserving the user experience can be significant. Currently, while image intent queries account for 25\% of the search queries, they go mostly unmonetized. Admittedly, image queries might not have the same commercial interest as web queries. However, even a small improvement in monetizing image queries can lead to tens of millions of dollars in incremental revenue. 

\head{Selected advertisement rejection}.  In general, we do not want
our approach to have a parameter (\eg a threshold) for
accepting/rejecting the ad image based on visual appearance. We choose
to keep it essentially \emph{parameter free}, by only using
consistency checks like reciprocity (see
Section~\ref{subsec:selection}) or local density (see clustering-based
placement approach in \S~\ref{subsec:place}) for possible
rejection of the selected ad; the Mean Shift algorithm does require a
scale parameter, that can however be estimated on-the-fly based on the
scale of the t-SNE output. To that end, we assume that the relevant
advertisement set $\cA$ has already been filtered by any other sense
(\eg financial viability, semantic similarity or user relevance) and
only includes high-quality ad candidates.  Additional parameter free
conditions can be considered, \eg rejecting the ad if it is on the
convex hull of the 2d t-SNE projection, but we found such further
checks unnecessary in practice.

\head{Diversification of the result set}. A by-product of our clustering-based positioning approach could be a way of diversifying the result set. Given that we calculate a density-based visual clustering of the top ranked images, one may choose to limit the number of images that are shown per cluster, therefore keeping the result set more visually diverse. Of course this implies that we analyze more images (\eg the first two pages of results) and we either completely drop some of them, or just place them further down in the results.

\head{Novelty analysis}. For the ad selection part, we use simple nearest neighbor search in feature space with compact visual features as in~\cite{BSCL14} and simple aggregation for extending nearest neighbor search to vector-to-set (dis-)similarity. 
% Although the aforementioned approach contains no technical novelty, to
To the best of our knowledge, this is the first paper that uses visual similarity for advertisement retrieval given a set of images. As we discuss before, the financial implications for such an approach are significant.
Further novelty comes from the use of two-dimensional embedding and density based clustering for presentation and ad placement, as well as from the use of state-of-the-art compact coding~\cite{KaAv14} on top of the compact feature vectors that enables our approach to run in a web-scale scenario. 
Finally, probably the most important contribution of this paper is the findings and implications that come from the analysis of our large crowd sourced study (see \S~\ref{subsec:exp_amt}). To the best of our knowledge, this is the first work that quantifies the effects that visual congruence  has to image ad placement and to users.

\head{Beyond image search}. In this paper we present our approach using image search as a target application. However, the VCAP algorithm can be applied in a larger pool of applications, ranging from personal photo galleries, to digital magazines. In the most extreme case it could be applied in any page that contains at least one image and we also want to place an advertisement.

%%% Local Variables:
%%% mode: latex
%%% TeX-master: "../www16"
%%% End:

%% file: tex/conclusions.tex
\section{Conclusions \& Future Work}
\label{sec:conclusions}
\vspace{.2cm}

In this paper we present an approach for placing native image-based ads in image search results. Our large crowd sourced user study shows that the proposed algorithm had a sizable and statistically significant impact on reducing user distraction. Our approach is able to retain brand recognition and awareness as effectively as a ``distracting'' ad would, while at the same time providing for a better user experience. Our method is analogous to print publications who utilized in-house design studios to ensure ads matched the magazine's content~\cite{vf:ads}, and demonstrates a similar application and utility online. 

We plan to run randomized field experiments~\cite{GeGr12} to measure the impact of our algorithm on the immediate and future user engagement.  Although we assume that only relevant ads are considered for placement by our approach, this is a restriction that may be relaxed if we want to further maximize visual similarity. It would be very interesting to see how the user experience changes in this setting, \ie when a user is presented with an ad that is not relevant to the images he sees semantically, but visually congruent.

In the future we also plan on investigating more into how visual congruence changes the visual perception of relevance for a user. As we found, it seems that users respond to ad relevance more if it is  blending visually with the result set. Delving deep into the result set and ad positioning is another interesting future direction, as well as diversification of results using the proposed ad placement algorithm. Of high practical importance is also to extend our space model and incorporate financial factors directly in our ad selection process.

%%% Local Variables:
%%% mode: latex
%%% TeX-master: "../www16"
%%% End: